# OBJECT VIEWPOINT CLASSIFICATION BASED 3D BOUNDING BOX ESTIMATION FOR AUTONOMOUS VEHICLES

*Zhou Lingtao, Fang Jiaojiao, Liu Guizhong*

School of Electronic and Information Engineering Xi'an Jiaotong University, Xi'an, China 710049
zhoulingtao7458@hotmail.com, 995541569@qq.com, liugz@xjtu.edu.cn

## ABSTRACT

3D object detection is one of the most important tasks for the perception systems of autonomous vehicles. With the significant success in the field of 2D object detection, several monocular image based 3D object detection algorithms have been proposed based on advanced 2D object detectors and the geometric constraints between the 2D and 3D bounding boxes. In this paper, we propose a novel method for determining the configuration of the 2D-3D geometric constraints which is based on the well-known 2D-3D two stage object detection framework. First, we discrete viewpoints in which the camera shots the object into 16 categories with respect to the observation relationship between camera and objects. Second, we design a viewpoint classifier by integrated a new sub-branch into the existing multi-branches CNN. Then, the configuration of geometric constraint between the 2D and 3D bounding boxes can be determined according to the output of this classifier. Extensive experiments on the KITTI dataset show that, our method not only improves the computational efficiency, but also increases the overall precision of the model, especially to the orientation angle estimation.

## 1. INTRODUCTION

While deep learning based 2D object detection algorithms [1, 2, 3, 4] have developed rapidly and achieved better and better robustness, monocular based 3D object detection remains a tricky problem. 3D object detection is a task to recover the 6 Degree of Freedom (DoF) poses and dimensions of objects in physical world which are the indispensable information for intelligent agents like autonomous vehicles to perceive and interact with the real world. Lacking depth information, image-based 3D object detection can be difficult due to the illness of projecting pixels on the image plane to 3D world coordinates which is eventually enabled via deep learning method [9]. Using a large amount of labelled samples, data driven CNN models can learn the empirical regulation between objects' appearance and their 3D properties with some scenario-specific priors.

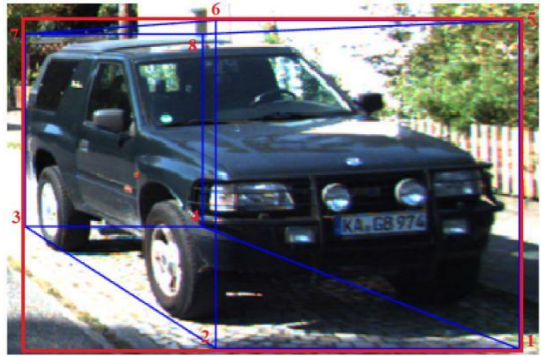

**Figure 1.** A sample from KITTI dataset illustrating relation between 2D and 3D bounding boxes, where the Arabic numbers indicate the indexes of the vertexes of the 3D bounding box.

In this work, we modify a state-of-the-art algorithm. In this method, the dimensions and orientations of objects are regressed using object-contained image patch produced by advanced 2D detector through a multi-branches CNN model and then locations of objects are determined via geometric constraints between 3D and 2D bounding boxes. To simplify the location inferring process, we split the viewpoints from which the objects are observed by camera into 16 categories and modify the existing multi-branches deep convolutional neural network by integrating a new appearance-related branch of viewpoint classification with the former CNN model to do this classification. And then the constraint configurations between 2D and 3D bounding boxes of the objects are determined according to object's viewpoint classification result.

The main contributions of this paper are three aspects: 1) a novel viewpoint classification method for determining the geometric constraint configuration being proposed to infer the 3D positions of objects; 2) a multi-branches CNN structure that simultaneously estimates dimensions, orientation and viewpoints classification; 3) an experimental evaluation on KITTI dataset demonstrating the improvement of our work on orientation estimation accuracy and overall detection precision compared with the baseline method.

## 2. RELATED WORK

A few methods have been proposed to address the problem of estimating 3D bounding boxes of objects by monocular RGB images data collected from driving scenarios. Taking advantage of the prior information of KITTI dataset, in [5], 3D sliding window and a sophisticated proposal method considering context on image, shape, location, and segmentation feature, is employed to generate 3D candidate boxes from scenes. The Fast R-CNN based detector is fed with image patches correspond to each 3D candidates to classify them and regress the bounding boxes and orientations of objects. [6, 7] separately detect objects that fall in different sub-categories in physical world. The sub-categories are defined by objects' shape, viewpoint and occlusion patterns and divided by clustering using 3D CAD object models. [10] proposes a method called deep MANTA which deploys a cascaded Faster-RCNN framework to detect objects and their parts on images from coarse-to-fine. 3D CAD models are also used to establish a library of templates to be matched with model's output so that the orientations and 3D positions of objects can be inferred. [8] proposes a 2D-3D object detection framework that regresses objects' orientations and dimensions from images patches containing objects utilizing 2D bounding boxes produced by advanced 2D detector. 2D-3D boxes geometric constraints are then be found to calculate the 3D positions of objects. Although the ergodic procedure of finding the most fitted configuration of constraint can be done in parallel, it still leads to unnecessary computational and time consumption which limits the application of this method.

Another kind of 3D detection methods use Lidar data as the additional information to get more accurate 3D detection results. [11] proposes a method that uses Lidar data to create multi front view and bird eye view (BEV) of the scene and feed them separately into a two-stage detection network along with the RGB image frames. After proposals are obtained from each view, a multi-stages feature fusion network is deployed to get the 3D properties and classes of objects. In [12], 3D anchor grids are used to get object candidates. After the extracted CNN features of each candidate from both images and Lidar-based BEV maps are fused and scored, top scored proposals are then classified and dimension-refined by the second stage of the network. This kind of methods produce much better 3D AP scores due to the known depth information, but the cost of computing resources, expensive devices as well as the power consumption for capturing depth information is not bearable for some circumstances. And these methods are like the way human perceiving the 3D physical world, in which no particular depth information is acquired, either.

Stereo vision is also used in some 3D detection algorithms for its simulation to human binocular vision. In [13], stereo images are used to get better 3D proposal in physical world. Stereo images based HHA features [14] which encoding the depth information of the scene are also used as a stream of input to get better 3D bounding box regression. [15] proposes a stereo-extended faster R-CNN detection method in which region proposals are generated on both left and right images from stereo pairs through RPN and their results are associated. After the keypoints, viewpoints and object dimensions are estimated from stereo proposals, a refinement procedure is deployed via region-based photometric alignment to get better detection.

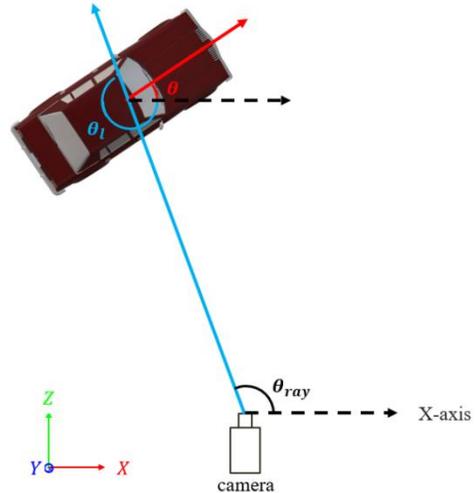

**Figure 2.** Illustration of local orientation $\theta_l$, angle of ray that from camera to a car's center $\theta_{ray}$ and the global orientation $\theta$ that needs to output.

## 3. 3D BOUNDING BOX ESTIMATION USING DEEP LEARNING AND GEOMETRY

Our work is based on the baseline method proposed in [8] which uses a two-stage 2D-3D detection procedure to infer 3D bounding boxes of objects. At the 2D detection stage, an advanced 2D detector is applied to determine the sizes and locations of objects on image plane. And then the cropped patches according to 2D detection results are fed into a multi-branches CNN to infer respectively: 1) objects' dimension; 2) objects' orientation, only the yaw angle to be specific in driving scenarios in KITTI dataset. Since the estimated dimensions and orientations of objects as well as the constraints between the projection of 3D boxes' vertexes and the edges of 2D detection results are given, the location of object can be recovered by solving an over-determined system of linear equations.

### 3.1 Correspondence Constraints

The fundamental idea of computing object's location comes from the consistency of 3D and 2D bounding boxes. In other words, the projection of the object's 3D bounding box which contains the object should fit tightly into the 2D detection window of the same detection. Four of eight vertexes of the 3D box should be projected right on the four edges of 2D window respectively. The cars showed in Fig. 1 is an actual sample from KITTI dataset, which shows one kind of correspondence configuration between 3D and 2D box: Vertex numbers 6, 2, 3, 1 are projected on the upper, lower, left, right edges respectively. Given the intrinsic matrix $K$ of camera, the 2D bounding box, $[u_{min}, u_{max}, v_{min}, v_{max}]$, the dimensions of object $d = [l, h, w, 1]^T$ and the global orientation $\theta$, these corresponding constraints can be formulated as:

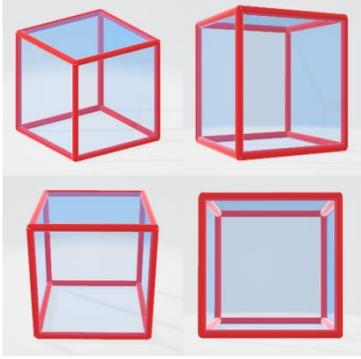

**Figure 3.** Illustration of 4 kinds of observational viewpoints.

$$\begin{cases} v_{min} = \pi_v(K[R_\theta \ T]S_3 d) \\ v_{max} = \pi_v(K[R_\theta \ T]S_4 d) \\ u_{min} = \pi_u(K[R_\theta \ T]S_1 d) \\ u_{max} = \pi_u(K[R_\theta \ T]S_2 d) \end{cases} \quad (1)$$

in which $R_\theta$ is the rotation matrix parameterized by orientation $\theta$. And $T = [t_x, t_y, t_z]^T$ denotes the transition from camera to the center of the bottom face of the object's 3D bounding box which needs to be solved from these equations.

$\pi_u$ and $\pi_v$ denote the image coordinates extracting functions getting the coordinates of object on the image plane:

$$\begin{aligned} \pi_u(P) &= p_1/p_3 \\ \pi_v(P) &= p_2/p_3 \\ P &= [p_1, p_2, p_3]^T \end{aligned} \quad (2)$$

And $S_1$ to $S_4$ are the vertexes selecting matrixes describing the positions of four selected vertexes using object's dimensions. These matrixes varied with different constraint configurations are used to define the relationship between 2D and 3D bounding boxes.

$$S_{1:4} = \begin{bmatrix} 0.5 & 0 & 0 & 0 \\ 0 & -1 & 0 & 0 \\ 0 & 0 & 0.5 & 0 \\ 0 & 0 & 0 & 1 \end{bmatrix}, \begin{bmatrix} 0.5 & 0 & 0 & 0 \\ 0 & 0 & 0 & 0 \\ 0 & 0 & -0.5 & 0 \\ 0 & 0 & 0 & 1 \end{bmatrix}, \\ \begin{bmatrix} -0.5 & 0 & 0 & 0 \\ 0 & 0 & 0 & 0 \\ 0 & 0 & -0.5 & 0 \\ 0 & 0 & 0 & 1 \end{bmatrix}, \begin{bmatrix} 0.5 & 0 & 0 & 0 \\ 0 & 0 & 0 & 0 \\ 0 & 0 & 0.5 & 0 \\ 0 & 0 & 0 & 1 \end{bmatrix} \quad (3)$$

Considering that object has zero pitch and roll angles in KITTI dataset, there can be a total of 64 kinds of correspondence constraint configurations that may occur according to the cases of the vertices projected correspondence to each side of the 2D boxes. For the car presented in Fig. 1.

[8] evaluates all the possible combination of vertex selecting matrixes to find the right one to calculate object's position. The key problem of this geometric constraint is to propose a simple and effective way for determining this corresponding configuration selecting matrixes.

### 3.2 Dimension and Orientation Regression

Due to the fact that object with same global orientation can look differently on image if their spatial position varies, [8] doesn't regress the global orientation $\theta$ of object directly. Instead, as showed in Fig. 2, a local orientation $\theta_l$ is estimated from image patch which is more dependent with the appearance of object. Then the global orientation we need can be computed as:

$$\theta = -(\theta_{ray} + \theta_l - 2\pi) = 2\pi - \theta_{ray} - \theta_l \quad (4)$$

where $\theta_{ray}$ denotes the angle between X-axis and ray from camera to object's center which can be computed easily using intrinsic camera parameters and object's location on image.

To avoid regressing an periodic angle value $\theta_l$, a MultiBin method that decomposing the orientation angle regression into the classification of angle's bins and regression of the residual between target angle and the center angle of each bin it falls in is used to get better angle estimation. The residuals of object's dimensions to the average size of each class are also regressed instead for the diversity of dimension distribution of objects in different classes.

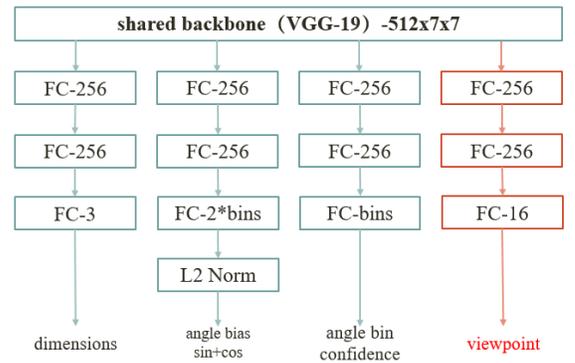

**Figure 4.** Architecture of our multi-branches network which consists four branches computing dimension residual, angle residual, confidence of each bin and classification of viewpoint respectively.

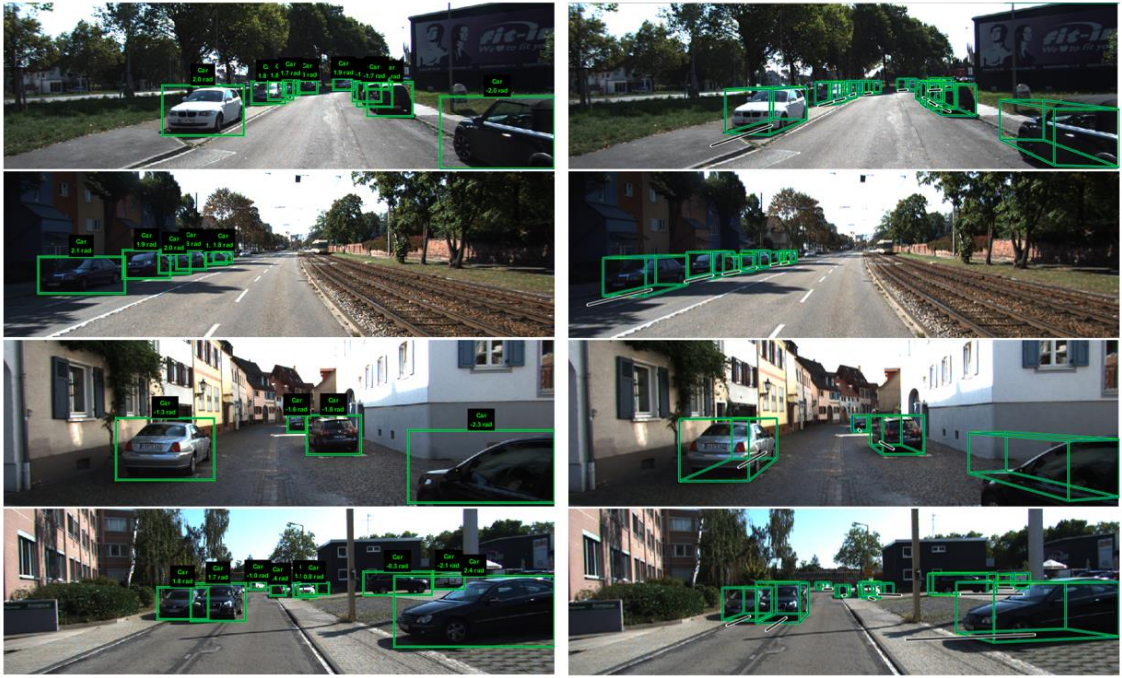

**Figure 5.** Result visualization of the 2D detection boxes (left, ported from [8]) and estimated 3D projections (right) for cars on KITTI dataset. The black lines attached to each 3D box represent the orientation of objects (start from the center of bottom face to the front of objects).

## 4. OBJECT VIEWPOINT CLASSIFICATION FOR CONSTRAINTS DETERMINATION

[8] exhaustively tests all possible correspondences to determine the location of object which is time and computational resource consuming. According to our observations, objects with the same constraint configuration have similar appearances on the image, which leads to the thought that setup a classification task for CNN to determine which configuration we should use to calculate objects' location according to the appearance information showed on image.

In spite of similarity of objects with same constraint configuration between their 2D and 3D boxes, objects with different constraint configuration can also share similar looks. For instance, the car showed in Fig. 1 can looks almost the same if its 3D bounding box's vertex number 5, 1, 3, 1 are prospectively projected on the upper, lower, left and right edge of its 2D window. To solve the too small and unbalanced distance between classes of configuration, we simply convert the classification of configuration to the classification of viewpoint from which camera observes object. As showed in Fig. 3, we have observed that there are four main classes of viewpoint in the image scenes of KITTI dataset [16]. In the vertical direction, camera can look down or front to objects. There're a small amount of cases that camera looks slightly up to objects, but we regard these objects as being looked from front view for convenience learning and the limited samples. In the horizontal direction, camera can observe only one side face or two of the 3d boxes.

Thus there're four combinations of two directions of views. Considering the orientation of objects, there're a total of 16 kinds of viewpoint. We merge several mostly similar constraint configurations to correspond with one kind of viewpoint, and therefore transfer the problem of determining correspondence configuration to estimate from which viewpoint object is observed. Note that by using this method for constraint determination, several kinds of configuration are merged into one, which can cause a little more but bearable error for localizing.

We let the viewpoint classification network share the backbone with other tasks as viewpoint classification can increase the model's sensitivity of object orientation. We believe that adding relevant task can lead to better network training. The complete network structure is showed in Fig. 4, we use an Image-Net pretrained VGG-19 network as backbone and add branches behind it to complete each task. Softmax loss is used to train both the viewpoint classification and the confidence of the MultiBin prediction. Dimension residual regression is trained by a simple L2 loss. As for angle residual regression, a L2-norm layer is added at the back of its branch to generate the sine and cosine prediction of residual angle and its loss is defined by cosine similarity between the prediction and the real residual angle:

$$L_{ang} = -\frac{1}{n_\theta} \Sigma \cos(\theta^* - c_i - \Delta\theta_i) \quad (5)$$

where $\theta^*$ denotes the ground-truth local orientation, $c_i$ denotes the $i$-th bin that ground-truth falls in, $\Delta\theta_i$ is the residual model predicts and $n_\theta$ is the number of overlapping bins covering the ground-truth.

| Method | Easy | | | Moderate | | | Hard | | |
|---|---|---|---|---|---|---|---|---|---|
| | AOS | AP | OS | AOS | AP | OS | AOS | AP | OS |
| Mono3D[5] | 91.01 | 92.33 | 0.9857 | 86.62 | 88.66 | 0.9769 | 76.84 | 78.96 | 0.9731 |
| SubCNN[7] | 90.67 | 90.81 | 0.9984 | 86.62 | 89.04 | 0.9952 | 78.68 | 79.27 | 0.9925 |
| 3D BBox[8] | 98.59 | 98.84 | 0.9974 | 96.69 | 97.20 | 0.9948 | 80.51 | 81.17 | 0.9919 |
| Ours | **98.79** | 98.84 | **0.9996** | **97.02** | 97.20 | **0.9980** | **80.80** | 81.17 | **0.9955** |

**Table 1.** Comparison of the Average Orientation Score(AOS, %), Average Precision(AP, %) and Orientation Score(OS) on KITTI dataset for cars.

| Method | Type | IoU=0.5 | | | IoU=0.7 | | |
|---|---|---|---|---|---|---|---|
| | | Easy | Moderate | Hard | Easy | Moderate | Hard |
| 3DOP[13] | stereo | 46.04 | 34.63 | 30.09 | 6.55 | 5.07 | 4.10 |
| Mono3D[5] | mono | 25.19 | 18.20 | 15.52 | 2.53 | 2.31 | 2.31 |
| 3D BBox[8] | mono | 27.04 | 20.55 | 15.88 | 5.85 | 4.10 | 3.84 |
| Ours | mono | **30.36** | **22.37** | **19.36** | **7.91** | **6.38** | **4.80** |

**Table 2.** Comparison of 3D Average Precision(%) on KITTI dataset for cars.

Thus the full loss function $L$ can be denoted as:

$$L = w_1 L_{dims} + w_2 L_{ang} + w_3 L_{conf} + w_4 L_{view} \quad (6)$$

$L_{dims}$, $L_{ang}$, $L_{conf}$ and $L_{view}$ denote loss for dimension regression, angle bias regression, confidence of bins and viewpoint classification respectively, while $w_{1:4}$ denote the weight factors of each loss.

## 5. EXPERIMENT

### 5.1. Implementation Details

Our 3D property estimation network was trained and tested on KITTI object dataset using the split used in [6]. We filtered out those samples which are heavily truncated from the training set excluding the potential harm to the model and randomly applied mirroring and color distortions to the training images for data augmentation. Then the network was trained with SGD at learning rate of 0.0001 for 20k iterations with a batch size of 8 to get the final weight used for validation. We set the weight factors of the losses $w_{1:4} = [1, 4, 8, 4]$. Fig. 5 shows some qualitative visualization of our result on KITTI validation set.

### 5.2. 3D Bounding Box Evaluation

We used the same 2D detection results as in [8] for making a fair comparison between our method and the baseline. Since many methods only released their result on cars, thus we make evaluation on the performance of our model on KITTI dataset for cars only.

**KITTI orientation accuracy.** Average Orientation Similarity (AOS), the official matric measuring orientation accuracy described in [1], is calculated to evaluate the performance of orientation estimation, as well as Orientation Score (OS) presented in [8]. OS equals AOS divided by Average Precision AP, which indicates the success of orientation estimation regardless of how well the performance of 2D localization. As shown in table 1, our method using the exact same 2D detector outperforms the baseline and other image based 3D detection methods.

**KITTI 3D bounding box metric.** 3D AP is used to evaluate the overall precision of 3D bounding boxes estimation. 3D Intersection-over-Union (IoU) of 0.7 and 0.5 are both used as threshold to determine a detection output is successful or not. Although monocular based 3D detection methods have many in spatial localization and therefore have low 3D AP score, as indicated in table 2, our method manages to perform relatively well compared to baseline and some other monocular based methods.

## 6. CONCLUSION AND FUTURE WORK

In this paper, we have proposed a method in which a viewpoint classification task is integrated to the 2D-3D two-stage detection framework to directly get the configuration between 3D and 2D bounding boxes fitting of object, based on which the location of object can be calculated.

Experiments demonstrate that our method not only is less time and computational resources consuming than the baseline algorithm which makes this method higher availability, but also performs better at estimating the orientation of objects through the joint training of viewpoint classification task and others. While our method achieves better performance than the baseline, it remains a problem that heavily truncated objects can't be localized well due to the assumption that 2D detection window and 3D box projection should always fit which is no more suitable for these kinds of objects which we hope to solve in the future.